\documentclass[a4paper,fleqn]{cas-dc}

\usepackage[numbers]{natbib}

\usepackage{graphicx}
\graphicspath{{figures/}}

\usepackage{amsmath,amssymb}

\usepackage{booktabs}
\usepackage{array}
\newcolumntype{Y}{>{\centering\arraybackslash}X}
\usepackage{tabularx}
\usepackage[flushleft]{threeparttable}
\usepackage{caption}
\usepackage{subcaption}
\usepackage{multirow}

\usepackage{pgfplots}
\pgfplotsset{compat=1.17}
\pgfplotsset{width=\linewidth}
\usetikzlibrary{matrix,calc}

\usepackage[linesnumbered,ruled,vlined]{algorithm2e}
\SetKwInput{KwInput}{Input}  
\SetKwInput{KwOutput}{Output}

\begin{document}

\let\WriteBookmarks\relax
\def\floatpagepagefraction{1}
\def\textpagefraction{.001}

\shorttitle{Lightweight Transformer in Federated Setting for Human Activity Recognition in Home Healthcare Applications}
\shortauthors{Raza et al.}  

\title[mode = title]{Lightweight Transformer in Federated Setting for Human Activity Recognition in Home Healthcare Applications}

\author[1,2]{Ali Raza}[orcid=0000-0001-8326-8325]
\cormark[1]
\ead{ali.raza@ensait.fr}
\credit{Conceptualization of this study, Methodology, Software, Data analysis, Paper writing}

\author[2]{Kim Phuc Tran}[orcid=0000-0002-6005-1497]
\cormark[1]
\ead{kim-phuc.tran@ensait.fr}
\credit{Conceptualization of this study, Methodology, Paper writing}

\author[2]{Ludovic Koehl}[orcid=0000-0002-3404-8462]
\ead{ludovic.koehl@ensait.fr}
\credit{Conceptualization of this study, Methodology, Paper writing}

\author[1]{Shujun Li}[orcid=0000-0001-5628-7328]
\cormark[1]
\ead{s.j.li@kent.ac.uk}
\ead[url]{http://www.hooklee.com/}
\credit{Conceptualization of this study, Methodology, Paper writing}

\author[2]{Xianyi Zeng}[orcid=0000-0002-3236-6766]
\ead{xianyi.zeng@ensait.fr}
\credit{Conceptualization of this study, Methodology, Paper writing}

\author[2]{Khaled Benzaidi}
\ead{khaled.benzaidi@ensait.fr}
\credit{Conceptualization of this study, Methodology, Data collection}

\author[3]{Sarah Hotham}
\ead{S.Hotham@kent.ac.uk}
\credit{Providing domain knowledge, Paper writing}

\affiliation[1]{organization={School of Computing \& Institute of Cyber Security for Society (iCSS), University of Kent},
city={Canterbury},
postcode={CT2 7NP},
country={UK}}
\affiliation[2]{organization={University of Lille, ENSAIT, GEMTEX--Laboratoire de Génie et Matériaux Textiles},
city={Lille},
postcode={F-59000},
country={France}}
\affiliation[3]{organization={Centre for Health Services Studies (CHSS), School of Social Policy, Sociology and Social Research (SSPSSR), University of Kent},
city={Canterbury},
postcode={CT2 7NF},
country={UK}}

\cortext[1]{Corresponding co-authors}



\begin{abstract}
Human activity recognition (HAR) is a machine learning task with important applications in healthcare especially in the context of home care of patients and older adults. HAR is often based on data collected from smart sensors, particularly smart home IoT devices such as smartphones, wearables and other body sensors. Deep learning techniques like convolutional neural networks (CNNs) and recurrent neural networks (RNNs) have been used for HAR, both in centralized and federated settings. However, these techniques have certain limitations: RNNs cannot be easily parallelized, CNNs have the limitation of sequence length, and both are computationally expensive. Moreover, in home healthcare applications the centralized approach can raise serious privacy concerns since the sensors used by a HAR classifier collect a lot of highly personal and sensitive data about people in the home. In this paper, to address some of such challenges facing HAR, we propose a novel lightweight (one-patch) transformer, which can combine the advantages of RNNs and CNNs without their major limitations, and also TransFed, a more privacy-friendly, federated learning-based HAR classifier using our proposed lightweight transformer. We designed a testbed to construct a new HAR dataset from five recruited human participants, and used the new dataset to evaluate the performance of the proposed HAR classifier in both federated and centralized settings. Additionally, we use another public dataset to evaluate the performance of the proposed HAR classifier in centralized setting to compare it with existing HAR classifiers. The experimental results showed that our proposed new solution outperformed state-of-the-art HAR classifiers based on CNNs and RNNs, whiling being more computationally efficient.
\end{abstract}


\begin{highlights}
\item We proposed a novel lightweight transformer for HAR classification in home healthcare applications. We show that the proposed transformer outperformed other state-of-the-art HAR classification methods based on CNNs and RNNs when trained and tested on a public dataset as well as a dataset we constructed.

\item In order to address challenges related to privacy and communication costs, we introduce TransFed, the first HAR classification framework based on federated learning and transformers.

\item We designed a prototype to collect human activity data using three different types of body sensors: accelerometer, gyroscope and magnetometer. We also tested different locations of each type of sensors on the human body to find the points of maximum impulse (PMIs) and evaluated the performance of the data for each location. We then constructed a new dataset for evaluating HAR classifiers, which will be released publicly as a new research dataset.

\item We evaluated the performance of TransFed on non-identical independent distributed (non-IID) data. Based on the data distribution we analyze how clients can affect the performance of TransFed.
\end{highlights}


\maketitle

\section{Introduction}
\label{sec:introduction}

Human activity recognition (HAR) is a classification task to learn which activity is performed by a certain person in a given period of time, which is normally achieved using supervised machine learning. Activities can be of different kinds such as sitting, standing, walking, running, eating, going upstairs and downstairs in a home environment. The rapid development of mobile computing, smart sensing and IoT technologies has led to a rich set of health-related data that can be used for various healthcare applications including HAR classifiers~\citep{Adnan2020unstructured_health_data}. For HAR, sensors such as wireless cameras, accelerometers, gyroscope sensors, wearables and other body sensors are often used.

HAR has important implications across a wide range of healthcare settings and contexts. For instance, it has a role in the prevention of diseases and maintenance of health (e.g., for monitoring rehabilitation, fall identification and prevention, tracking health behaviors) and clinician- and patient-focused tools (e.g., point-of-care diagnostics and early detection of diseases) that facilitate management of health conditions and prompt behavioral changes with real-time feedback~\citep{WuLuo2019wearable_in_healthcare}. It can also be used for fitness and exercise monitoring, e.g., used by mobile apps on smartphones and wearables that count steps and be context-aware, and for safety purposes, e.g., the ``Do Not Disturb while Driving'' feature on iOS version 11~\citep{applesupport_2021}, which requires real-time detection of the specific human activity ``driving''.

Recently, by utilizing deep learning, researchers have made substantial progress on high-accuracy HAR classifiers~\citep{yao2017deepsense, ordonez2016deep, hammerla2016deep}. For instance,  Yao et al.~\citep{yao2017deepsense} proposed a deep learning model based on data collected from accelerometer and gyroscope sensors on mobile devices, which involves a hybrid model combining a convolutional neural network (CNN) and a recurrent neural network (RNN). However, training such models on real-world data collected from smart devices leads to two major challenges. First, deep learning requires a large amount of data for training~\citep{yao2017deepsense}, which will incur communications between the centralized server and clients (data owners). Second, collecting data from a home environment can raise privacy concerns since a lot of the sensor data include or can infer personal and sensitive data about people in the home~\citep{liu2020privacy, de2021critical, Char2018ML_in_healthcare}. Third, even through for healthcare applications there are often a lot of data and sensors available, the relevant data owners and local healthcare providers may not be willing to share the data for legitimate business interests, therefore limiting the data available for training HAR classifiers.

The above challenges can be addressed by a new machine learning concept called federated learning~\cite{li2020federated}, which allows multiple clients to collaboratively build a global model without sharing their local data with each other or with a global server, but by sharing only trained parameters of local models. This helps improve privacy of local data, reduce unnecessary communications between the global server and the clients, and meets the business needs of data owners and local healthcare providers who would not share their data. Therefore, many researchers have considered using federated learning for HAR and other healthcare applications~\citep{xu2021federated}. For instance, Liu et al.~\citep{liu2020fedsel} developed a classifier in a federated setting to address the aforementioned challenges by training a classifier using federated learning~\citep{bonawitz2019towards}, and Sozinov et al.~\cite{sozinov2018human} compared HAR classifier trained using centralized learning with HAR classifier trained using federated learning for different data distributions among clients. Their results show that federated learning based classifiers achieved a comparable accuracy to deep learning classifiers in a centralized setting.

In regards to performance, commonly used deep learning techniques including CNNs and RNNs have their limitations. CNNs have an advantage over RNNs (including LSTMs) as they are easy to parallelize, while RNNs have recurrent connections and hence parallelizaion cannot be easily achieved. However, in long-term sequences like time-series, capturing the dependencies can be cumbersome and unpractical using CNNs~\citep{yin2017comparative}. To address these challenges, transformers have been introduced recently~\citep{vaswani2017attention}. The transformer technique is an attempt to capture the best of both worlds (CNNs and RNNs). They can model dependencies over the whole range of a sequence and there are no recurrent connections, so the whole model can be computed in a very efficient feedforward fashion. Since its introduction, transformers have been widely studied in various applications, for example, in natural language processing~\citep{wolf2020transformers} and healthcare~\citep{yang2020clinical}. However, transformers are not developed for HAR, and developing transformers in federated setting can potentially boost up the HAR. Nevertheless, there are open questions of the implementation details such as what are the limits of the algorithm and if transformers really perform well in HAR, coupled with federated learning.

In this study, to answer all these questions, first we develop a novel lightweight transformer for HAR classification and showed that it can provide high performance in terms of accuracy and computational cost compared to existing deep models such as RNNs and CNNs. We trained and tested the proposed base-line transformer using different open-source sensor data, as well as on some data that we collected using a prototype that we developed to collect the human activity data using three different types of sensors: accelerometer, gyroscope and magnetometer. While collecting the data we tested different sensor locations on human body: hip, chest and upper arm (further details about the prototype and data collected will be provided in later sections). Furthermore, to address the aforementioned challenges, such as privacy concerns and additional communication costs, we propose a novel transformer framework in federated setting called TransFed, the first transformer-based classifier for HAR in federated settings. Moreover, we evaluated performance of federated learning using the proposed transformer and showed that federated learning can be used instead of centralized learning for HAR classifier. We compared the proposed transformer-based HAR classifier under two training settings, using centralized learning with HAR classifier and using federated learning. In federated learning, we use skewed data based on a non-identical independent distributed (non-IID) data distribution among clients, and in the centralized setting we use the whole dataset which contains only slightly unbalanced classes. The results showed that the proposed FedTrans outperformed existing state-of-the-art methods.

The main contributions of the paper are as follows:
\begin{enumerate}
\item We proposed a novel lightweight transformer for HAR classification. We show that the proposed transformer outperformed other state-of-the-art HAR classification methods based on CNNs and RNNs when trained and tested on a public dataset as well as a dataset we constructed.

\item In order to address challenges related to privacy and communication costs, we introduce TransFed, the first framework for HAR classification based on federated learning and transformers. 

\item We designed a prototype to collected human activity data using three different types of body sensors: accelerometer, gyroscope and magnetometer. We also tested different locations of each type of sensors on the human body to find the points of maximum impulse (PMIs) and evaluated the performance of the data for each location.  We then constructed a new dataset for evaluating HAR classifiers, which will be released publicly as a new research dataset.

\item We evaluated the performance of TransFed using non-identical independent distributed (non-IID) data. Based on the data distribution we analyze how the non-IID data of clients can affect the performance of TransFed.
\end{enumerate}

The rest of the paper is organized as follows. In the next section, we briefly introduce some related work on HAR, federated learning and transformers. Section~\ref{section:proposed_methods} explains our proposed transformer and the federated learning framework TransFed in details. The experimental setup and the experimental results on performance analysis are given in Sections~\ref{section:experimental_setup} and \ref{section:performance}, respectively. Some further discussions including limitations and some future research directions are presented in Section~\ref{section:discussions}. The last section concludes the paper.

\section{Related Work}
\label{section:related_work}

\subsection{Human Activity Recognition}

Based on data from one or more body sensors, HAR classification using a classifier to predict human activities. Generally, the data contains tri-axial data from different sensors like accelerometer, gyroscope and magnetometer. Most modern smart devices such as smartphones and wearables have such sensors. Initially, researchers have used various hand-crafted features for training HAR classifiers.

Generally speaking, the hand-crafted features can be divided into three main types: frequency domain, time domain, and time-frequency analysis. Classical machine learning models like $k$-means, probabilistic methods (naive Bayes) and support vector machines have been proposed for HAR classification~\citep{mannini2010machine}.
 
\begin{figure}[!ht]
\centering
\includegraphics[width=0.4\linewidth]{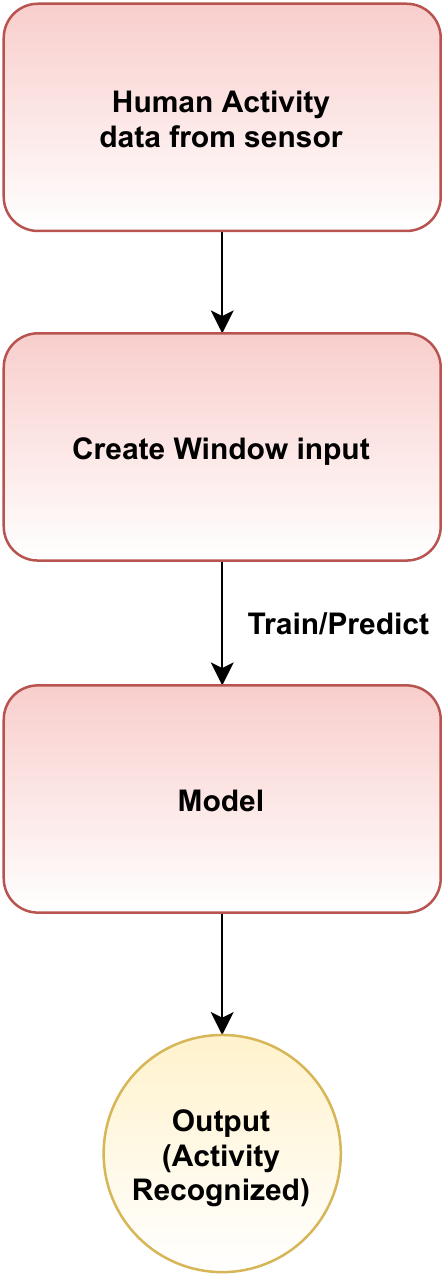}
\caption{Overview of the HAR classification pipeline.}
\label{fig:overview_HAR_classification}
\end{figure}

On the other hand, deep learning exploits benefits of having a huge amount of data and highly non-linear deep models, to outperform classical models. Using deep learning, the feature extraction simply can be omitted which in case of classical machine learning is a hectic and important task. Raw data in the form of sliding window or simple window can be directly fed into deep learning based classifiers. CNNs, long short-term memory (LTSM) RNNs, and hybrid models combining RNNs and CNNs are dominating approaches proposed for HAR~\citep{ChenYang2015SMC, yao2017deepsense, attal2015physical, sikder2021ku, sozinov2018human, hassan2018robust, ignatov2018real, murad2017deep, Zhou2020}. Figure~\ref{fig:overview_HAR_classification} shows the pipeline of a typical HAR classification process. First, the raw data from sensors is transformed into windows of a fixed-length size, which are fed directly to the classifier. While prediction, data is collected using the same window length and then, again depending on the model selected for a HAR classifier, features are extracted or raw windowed data is fed into the classifier that predicts the target human activity, such as walking, sitting, etc.

\subsection{Federated Learning}
 
Federated learning (FL)~\citep{mcmahan2016federated} is a new concept of distributed machine learning, where peers in a network train a global model collaboratively, without sharing their local training data with the central node. The main goal of FL is to train a global model $\mathbf{GM}$ by using shared parameters of locally trained models at $K>1$ clients. This ensures that the local training data remains at each client. At the $r$-th global round, each client $P_k$ has it own local data $D_k^r$ and trains a local model $\mathbf{LM}_k^r$. After training, each client sends the parameter of $\mathbf{LM}_k^r$ to the central server, which aggregates the local parameters from all clients to produce an updated global model $\mathbf{GM}^r$. There are mainly two different approaches for making global updates: i) federated averaging where clients send the updates to global server after training the local model for multiple training epochs, and ii) federated stochastic gradient descent (SGD) where clients send updates to the global server after each local training batch. McMahan et al.~\cite{mcmahan2016federated} compared the two approaches and shown that federated averaging can reduce communication costs by a factor of 10 to 100 times, compared to federated SGD. The merits of federated averaging make it popular in many applications~\citep{yang2019federated}. Federated averaging can be described by the following equation:
\begin{equation}
\mathbf{LM}_k^{r+1} = \mathbf{LM}_k^r-\alpha g_k^r;~\mathbf{GM}^{r+1}=\sum_{k=1}^{K}\frac{n_k}{n} \mathbf{LM}_k^{r+1},
\end{equation}
where $g_k^r$ is the gradient of back-propagation at the $r$-th global round, $\alpha$ is the learning rate, $n$ is the sum of data points of all the participating clients, and $n_k$ is the number of data points of the $k$-th client.

Federated learning has many advantages over the centralized approach. For instance, one of its advantages over the centralized approach is that it can provide more privacy protection to the sensitive data. This is because the global model is trained without requiring clients to share their local (often sensitive) data. Moreover, it reduces the communication costs because only trained parameters are shared, instead of the often large amount of data from all clients.

\subsection{Transformers}

\begin{figure*}[!ht]
\centering
\includegraphics[width=0.8\linewidth]{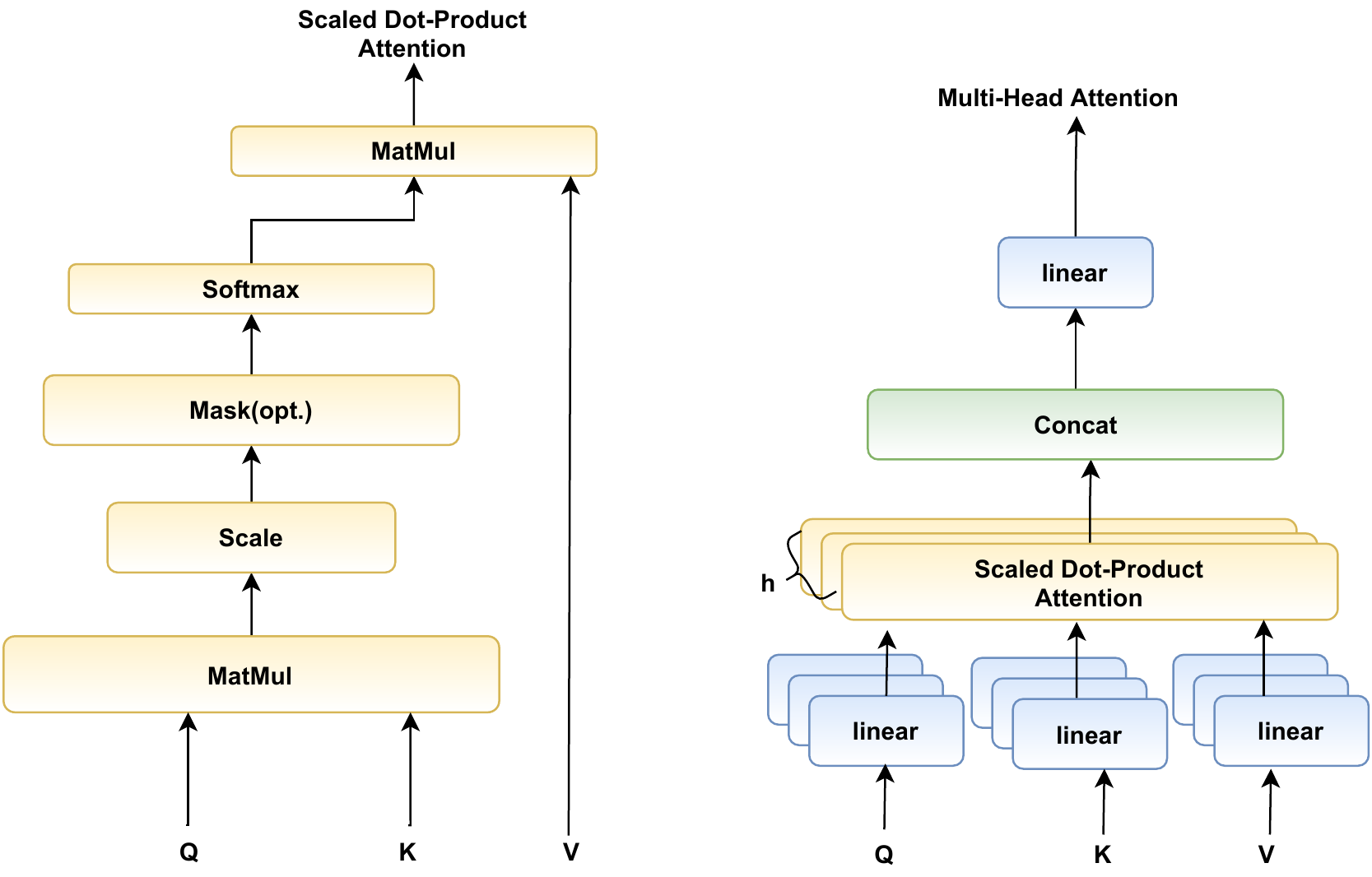}
\caption{An illustrative diagram of the  attention mechanism and its parallelization, regenerated from~\citep{vaswani2017attention}. Left: the attention mechanism, Right: parallelization of the attention mechanism.}
\label{fig:attention_mechanism}
\end{figure*}

Vaswani et al.~\cite{vaswani2017attention} introduced a novel architecture called Transformer for sequence to sequence learning. One of the key component of transformers is the attention-mechanism. The attention-mechanism looks at an input sequence and decides at each step which other parts of the sequence are important. Similar to LSTMs, a transformer basically transforms one sequence to another one with the help of two parts: an encoder and a decoder, but it differs from existing sequence-to-sequence methods in that it does not imply any recurrent networks (GRU, LSTM, etc.). The encoder and decoder consist of modules that can be stacked on the top of each other multiples times. Each module mainly consists of multi-head attention and feed-forward layers. The input and output are first embedded into an $n$-dimensional space since they cannot be used directly. Another part of the model is positional encoding of different words. Since there are not recurrent networks that can remember how a sequence is fed into the model, a relative position is encoded for every part of the input sequence. This positions are added to the embedded n-dimensional vector of each input sub-sequence.

The attention mechanism used in transformers can be described by the following equation:
\begin{equation}
\text{Attention}(Q,K,V) = \text{Sofmax}\left(\frac{QK^T}{\sqrt{d_k}}\right)V,
\end{equation}
where $Q$ is the query matrix (vector representation of input sub-sequence), $K$ are all the keys (vector representations of all the sequence) and $V$ are the values(vector representations of all the sequence). For the encoder and the decoder, multi-head attention modules, $V$ consists of the same word sequence as $Q$. However, for the attention module that is taking in the encoder and the decoder sequences, $V$ is different from the sequence represented by $Q$. In other words, $V$ is multiplied and summed with the attention weights $\gamma$, defined by the following equation:
\begin{equation}
\gamma = \text{Sofmax}\left(\frac{QK^T}{\sqrt{d_k}}\right).
\end{equation}
The self-attention mechanism is applied multiples times in parallel along with the linear projections of $Q$, $K$ and $V$. It helps the system to learn from different representations of $Q$, $K$ and $V$. The weight matrices $W$ that are learned during the training are multiplied by $Q$, $K$ and $V$ to learn the linear representations. Figure~\ref{fig:attention_mechanism} gives an illustrative diagram of the attention mechanism and its parallelization.

Moreover, positional encoding is used to keep track of the input and output sequence. Finally, transformers employ feed-forward networks. These feed-forward networks have identical parameters for each position of the input sequence, which describes each element from a given sequence as a separate but identical linear transformation.

Transformers have various applications because they use attention mechanisms, which can help the machine learning models to learn from data more effectively to improve the performance of many machine learning tasks such as natural language processing ones~\citep{devlin2018bert}. For example, a hybrid HAR classifier using CNNs and transformers was introduced in~\citep{li2021two}, which utilizes a two-streamed structure to capture both time-over-channel and channel-over-time features, and use the multi-scale convolution augmented transformer to capture range-based patterns.

\section{Proposed Methods}
\label{section:proposed_methods}

In this section, we explain our proposed methods, including the proposed lightweight transformer and the FL-based framework TransFed for addressing privacy concerns in a federated setting. For the proposed transformer model, we first describe the model itself and then move on to explain two data formats we tested for the proposed transformer model. Both the model itself and the data formats tailored for the model help improve the transformer's performance in terms of computational complexity and classification accuracy. The following three subsections will introduce the transformer model, the data formats used, and the proposed TransFed framework, respectively.

\subsection{Proposed Lightweight Transformer Model}

\begin{figure}[!ht]
\centering
\includegraphics[width=\linewidth]{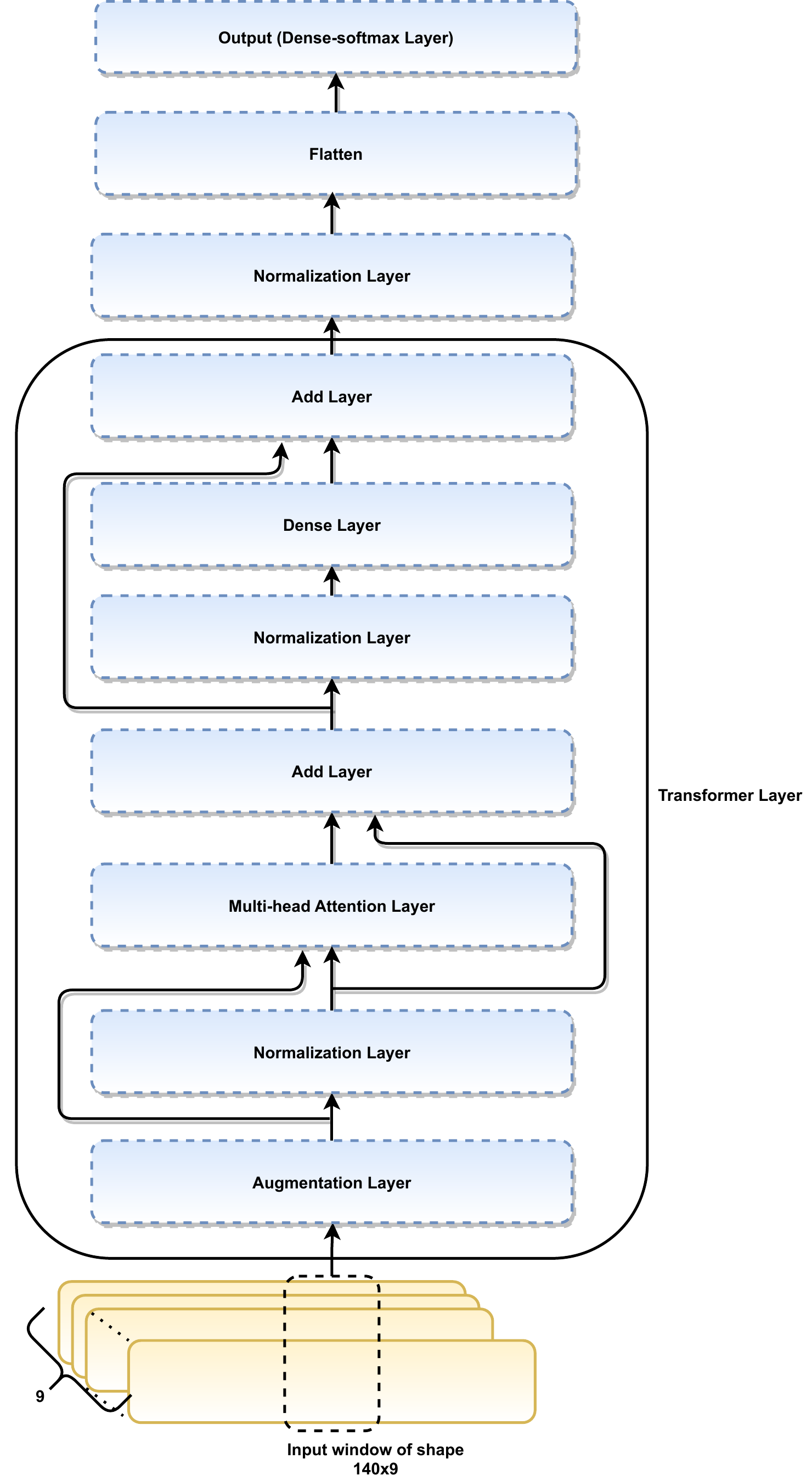}
\caption{The proposed transformer model.}
\label{fig:transformer}
\end{figure}

We designed a novel lightweight transformer, as shown in Figure~\ref{fig:transformer}. From bottom to top, the first layer inputs the raw human activity data after preprocessing it into certain sized window (as discussed earlier). The input is then passed through the transformer layer, which extracts discriminative features from the input data. Finally, the output of transformer is fed into a prediction layer for the classification or final output.

Unlike the traditional transformer model, we use a single patch encoding in our model. This is because we found out that, we can get the state-of-the-art results even without using multiple patches, which helps make our proposed transformer simpler and therefore more computationally efficient. The proposed transformer is virtually divided into two parts: the transformer layer and the prediction layer. Both layer are composed of many sub-layers. The transformer layer starts with an augmentation sub-layer, which is used to increase the diversity of the training set by applying random (but realistic) transformations. The output of the augmentation sub-layer is fed into a normalization sub-layer to normalize the data. Following the normalization sub-layer, a multi-head attention sub-layer applies the self-attention mechanism to the input data, which is the fundamental mechanism of a transformer. The self-attention mechanism is a sequence-to-sequence operation: a sequence of vectors goes in, and a sequence of vectors comes out. Let us call the input vectors $X_1, X_2, \dots, X_l$ and the corresponding output vectors $Y_1, Y_2, \dots, Y_l$. To produce the vector $Y_i$, the self-attention operation applies weights averaged over the input vectors as follows:
\begin{equation}
y_i=\sum_jw_{ij}X_j,
\end{equation}
where $j$ indexes over the whole sequence and the weights sum to one over all $j$. The weight $w_{ij}$ is not a parameter as in a normal neural network, but derived from a function over $X_i$ and $X_j$. To make this operation more lightweight, we use the dot product:
\begin{equation}
w_{ij}=X_i^TX_j.
\end{equation}
Since the dot product gives a value between $-\infty$ and $+\infty$, we apply a softmax to map the values to the range of 0 and 1, and to ensure that they sum to 1 over the whole sequence:
\begin{equation}
w_{ij}=\frac{\text{expw}_{ij}}{\sum_j\text{exp w}_{ij}}.
\end{equation}
The self-attention operation defines the correlation among the input features with respect to the learning task. The output of the multi-head attention sub-layer is then added with the output of the previous normalization sub-layer and fed the result into a succeeding normalization sub-layer, which is then passed through a dense sub-layer. The dense sub-layer applies a non-linear transformation for further feature extraction, given as:
\begin{equation}
\text{Output} = \text{activation} (\text{dot}(\text{input},\text{kernel})),
\end{equation}
where ``activation'' is the element-wise activation function passed as the activation argument, ``kernel'' is a weight matrix and ``dot'' is the dot product. The output of the dense sub-layer is added with the output of the previous addition sub-layer. The transformer layer can be applied multiple times. The output from the transformer layer is fed into a flattening and dense output sub-layer with softmax activation, which gives the final classification probability distribution over the pre-defined classes.

\subsection{Data Formats for Proposed Transformer Model}

Since the input data format plays an important role in the classification and computational efficiency of the proposed transformer model, we experimented with two possible formats of the input data as explained below.

\subsubsection{Image-based Format}

\begin{figure*}[!ht]
\centering
\includegraphics[width=0.9\linewidth]{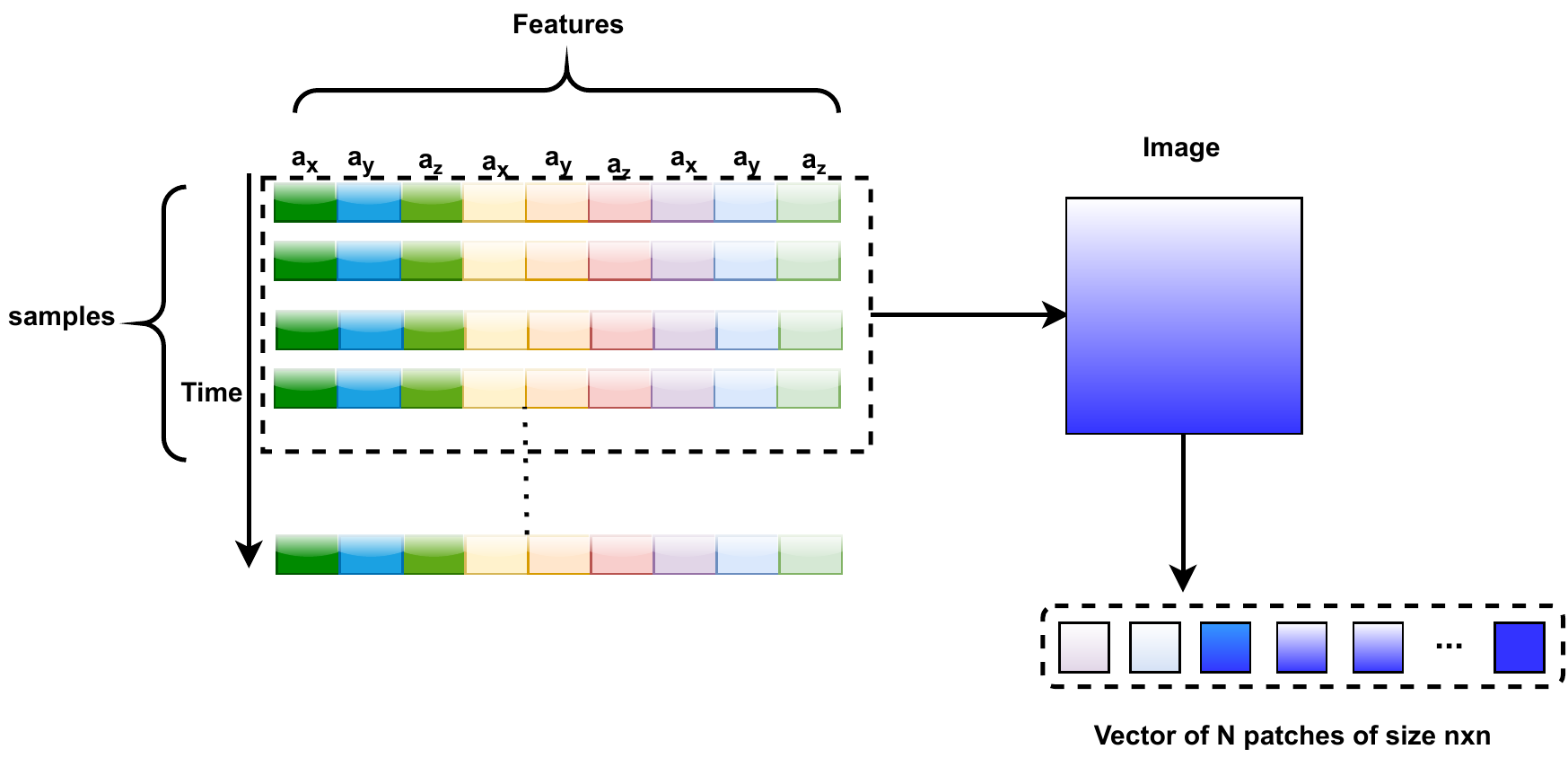}
\caption{Image-based format for the proposed transformer model.}
\label{fig:image_creation}
\end{figure*}

Since the transformers were initially introduce for computer visions tasks, they work with 2-D images by splitting an image into a vector of small sub-images (patches). This vector is used as the input. In order to follow the convention, we created an image of size $N\times M$ from input samples of an activity in a time frame of 2 seconds, as a time frame of 2 seconds can give enough information about the activity, as shown in Figure~\ref{fig:image_creation}. After this, patches of size $n \times n$ were created, where $N,M>n$. We tried images and patches of different sizes. Unfortunately, when we trained and tested the proposed transformer, using the above-mentioned data format, the results were not good enough. This is due to broken the natural boundary between consecutive 1-D samples into 2-D images, which makes it hard for the transformer to capture all features effectively by suppressing all random noise in the images because no any 2-D filter of a reasonable size would cover distant pixels that are neighboring samples in the original 1-D signal. We observed that with the increase in size (the number of transformer layers), the classification performance was improved but this setting is not suitable for edge devices with a relatively low computation power. Nevertheless, this approach can be utilized in settings where the computational power is high enough. Hence, we decided not to use the image-based format. 

\subsubsection{Averaged-Window Format}

\begin{figure*}[!ht]
\centering
\includegraphics[width=0.9\linewidth]{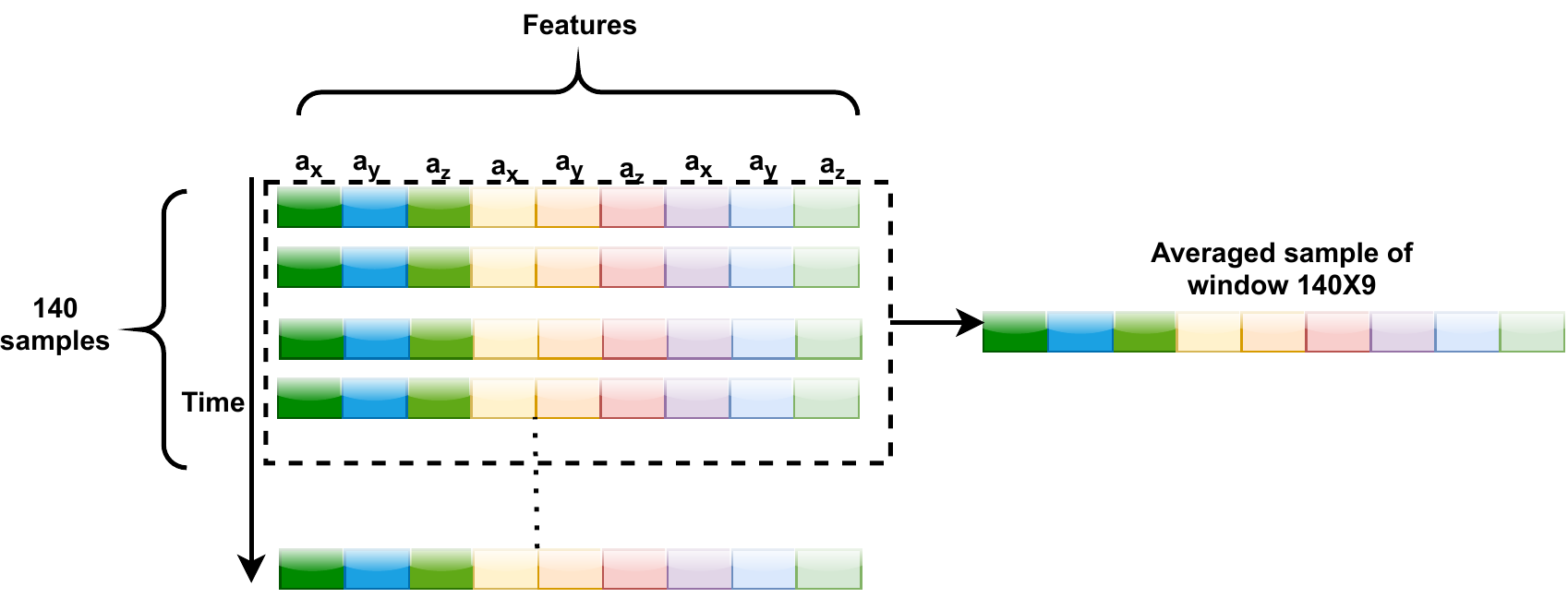}
\caption{Averaged-window format for the proposed transformer model.}
\label{fig:window_creation}
\end{figure*}

In this approach, we propose to reshape the raw input data into fixed sliding window of shape $W\times F$, where $W$ is the number of samples in a window and $F$ is the number of features. In our experiments, we tried to change $W$ as well as the number of features. Since a human activity is an action that can be performed in a certain window of time, to capture a specific activity we need to optimize the window size so that the model can map the information contained in that window to a specific class. For example, going upstairs is an activity that can be recognized within a suitable window size such as 2 or 3 seconds. Hence, $W$ (the window size) is very critical in HAR because of the following trade-off: if $W$ is too small the classifier cannot distinguish between activities; and if $W$ is too large the classifier will require more data and more computational resources. Let us take going upstairs as an example. If $W$ is below one second, it would not be suitable to recognize the activity as no human being can finish going upstairs within just one second (for HAR tasks, a single temporal point is too small to be informative). In regards to $F$, we noticed that when there are a large number of candidate classes, a large $F$ significantly improves the classification performance. While for a small number of classes, a small (such as 6) or large (such as 9) value of $F$ gives almost the same classification performance. Moreover, we found that combining tri-axial data from accelerometer and gyroscope provides a significant improvement over combining accelerometer and magnetometer data, and combining magnetometer and gyroscope data. After trying many different values of the window size with grid search, we found that a window size of $140\times 9$ provided optimized results for all human activities we considered, when both accuracy and computational costs are taken into account. To summarize, we take samples of the input over a time frame of 2 seconds with 9 features and average them feature-wise to generate a new sample as shown in Figure~\ref{fig:window_creation}. This format can work because averaging cancels the random noise.

\subsection{The Proposed FL Framework TransFed}

\begin{figure}[!ht]
\centering
\includegraphics[width=\linewidth]{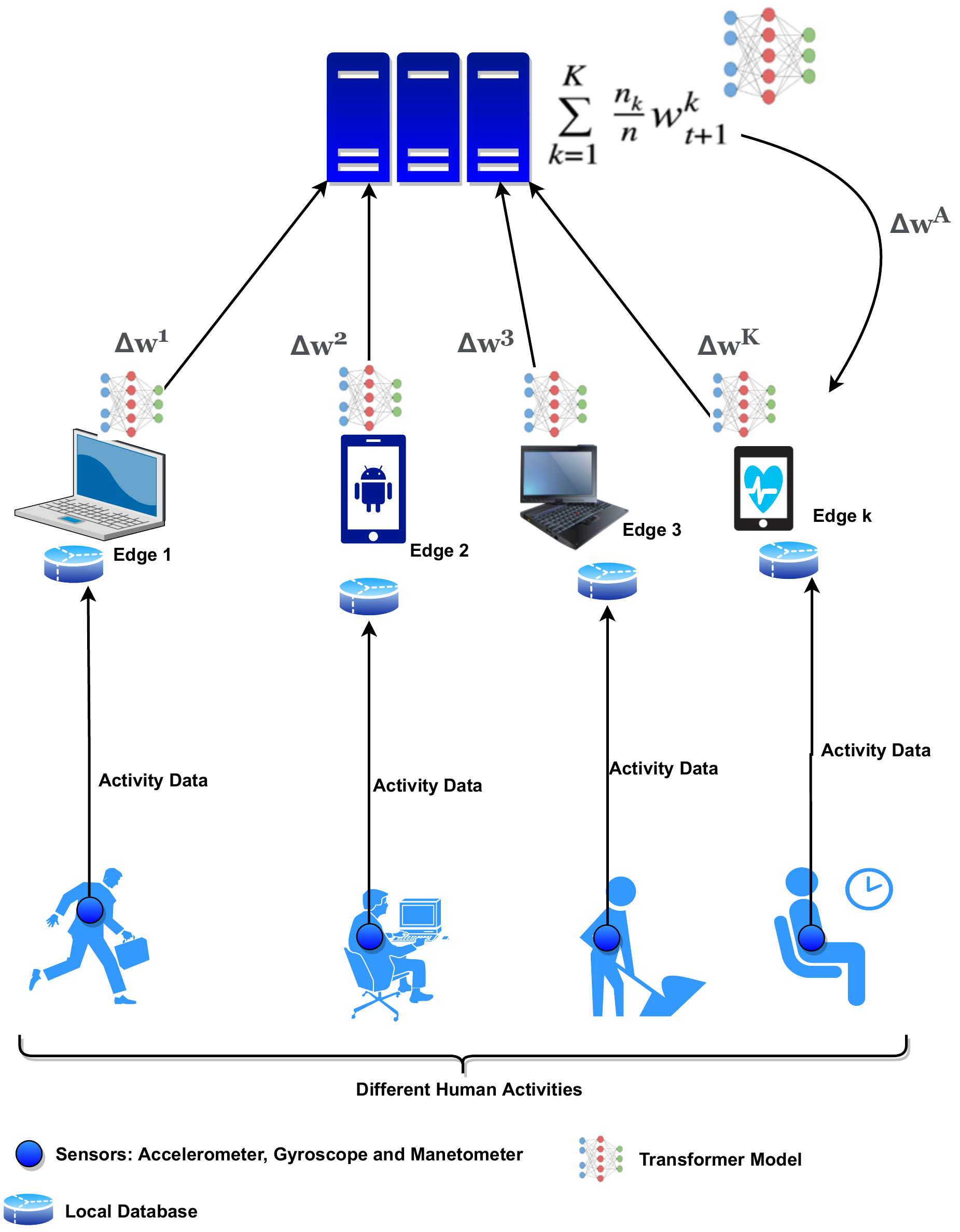}
\caption{An illustrative diagram of the proposed framework.}
\label{fig:overview}
\end{figure}

Figure~\ref{fig:overview} shows the basic flowchart of the proposed TransFed. In our method, the federated setting is adopted in order to facilitate collaborative learning while preserving the privacy of the underlying data. In the federated setting, a central (global) server sends the compiled architecture of the model (which is a transformer in our case) to all edge or client\footnote{In this paper, we use the terms `edge' and `client' interchangeably.} devices. Each devices trains its transformer locally using its local data. After all the local transformers are trained, each edge device sends trained parameters of its transformer to the global server, which are then aggregated by the global server to construct the global model without a training process. After the global model is available, each edge device downloads the aggregated parameters of the global model and updates its local model according to its local needs. There are two expected key advantages of the federated setting, (i) it increases the overall accuracy and the generalization of the model, and (ii) it provides better privacy protection to the data owners. Algorithm~\ref{algo:algorithm} defines the workflow of the proposed TransFed framework.

\section{Experimental Setup}
\label{section:experimental_setup}

In this section, we discuss the experimental setup that we designed to test the performance of the proposed transformer in federated setting using real-time data. We first discuss a testbed that we designed to construct a new dataset and to support the experiments for evaluating the performance of the proposed the TransFed framework.

\subsection{Testbed for Data Collection}

\begin{figure}[!ht]
\centering
\includegraphics[width=\linewidth]{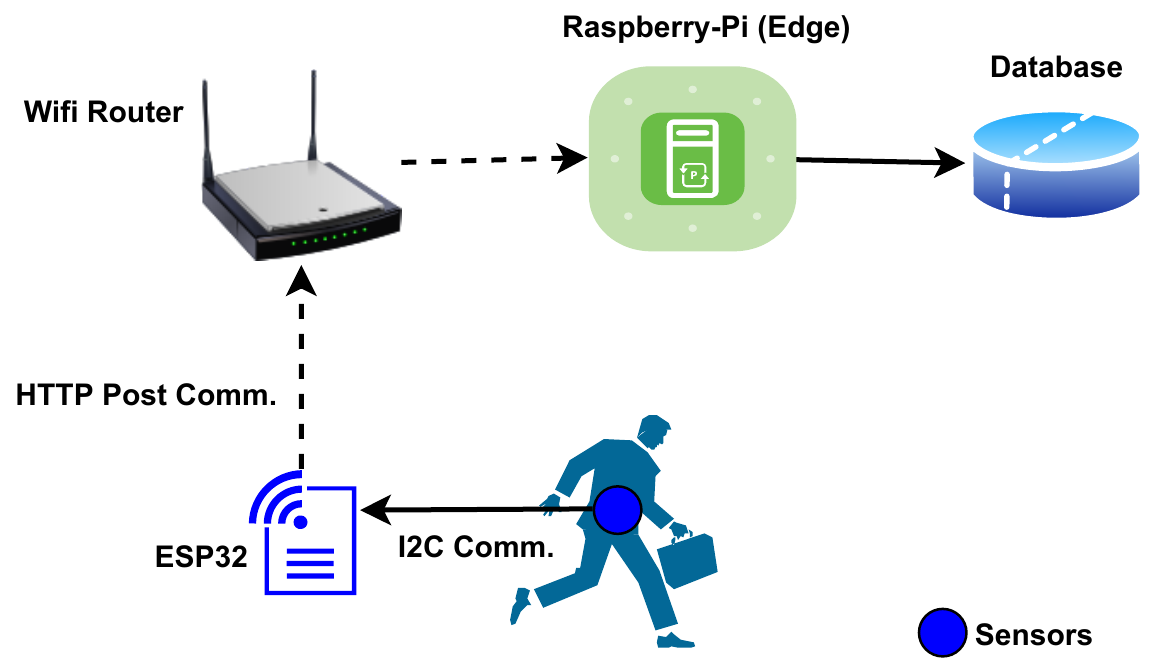}
\caption{Our prototype for real-time data collection.}
\label{fig:prototype}
\end{figure}

TransFed can in principle work with different types of sensors from which data about human activities are collected. For our testbed, we decided to use three types of sensors available on most smart wearable devices: tri-axial accelerometers, gyroscopes and magnetometers. These sensors provide measurements at a sampling frequency of 115 Hz. The frequency of 115 Hz establishes a sufficient condition for a sample rate that permits a discrete sequence of samples to capture all the information from a continuous-time human activity signal. The testbed is shown in Figure~\ref{fig:prototype}.

\subsubsection{Sensor Locations and Data Collection}

Since the quality of data being used by an ML model can significantly impact its performance, we decided to use a data-centric approach to ensure the performance of our proposed model. One important aspects of acquiring high-quality data for HAR purposes is to identify an optimized location on the human body for each sensor used. We need to optimize the sensor locations in such a way that it provides both (i) data informative enough to be used in ML, and (ii) convenience and comfort to humans while they are wearing the sensors. Keeping both points in mind and considering what was commonly chosen in the literature~\cite{sikder2021ku, Zhou2020}, we tried different locations on the human body: upper arm, hip and chest. After placing each sensor on each body location in the recruited individuals, we recorded data for the following 15 activities: sitting, walking, jogging, going upstairs, going downstairs, eating, writing, using laptop, washing face, washing hand, swiping, vacuuming, dusting a surface, and brushing teeth. Using the data collected from each location, we trained and tested the proposed transformer model, separately for the data of each location, in a centralized setting. The results showed that the model trained and tested using data collected from hip outperformed models using data collected from two other locations (chest and upper arm). Hence, in the following we report experimental results with data collected from hip only.

To construct our new dataset, we recruited five human participants with different ethnic backgrounds, i.e., Pakistani, Algerian, French, Vietnamese and Moroccan. Before collecting the data, each participant was briefed about each activity to be conducted, the health hazards of the experimental setup, and how the data will be used by the researchers. No financial compensations were made. Each participant performed the 15 activities as shown in Table~\ref{tab:dataset}. While performing each activity for a duration of around 3 minutes\footnote{The precise duration was determined based on the ability and willingness of each participant.}, the sensors placed on each participant generated activity data, which was sent to the ESP32 module using the I2C communication protocol. Note that the ESP32 and sensors were both located on the participant's body, powered by a lithium battery. The ESP32 module sent the data over Wi-Fi using the HTTP method POST to the edge (Raspberry Pi), which was then stored in a MySQL database. The experiment was approved by the ENSAIT, GEMTEX--Laboratoire de Génie et Matériaux Textiles, University of Lille, France, from which all participants were recruited.

\begin{table*}[!ht]
\centering
\begin{threeparttable}
\centering
\caption{Description of the 15 human activities covered in our experiments for constructing the new dataset.}
\label{tab:dataset}
\begin{tabularx}{\linewidth}{ccccY}
\toprule
Class Name & Class ID & Performed Activity & Number of Samples collected\\
\midrule
Standing & 0 & Standing still on the floor  & 22,851\\
Sitting & 1 & Sitting still on a chair  & 23,204\\
Walking & 2 & Walking at a normal pace  & 23,982\\
Jogging & 3 & Running at a high speed &  21,594\\
Going Upstairs & 4 & Ascending on a set of stairs at a normal pace  & 23,832\\
Going Downstairs & 5 & Descending from a set of stairs at a normal pace  & 21,836 \\
Eating & 6 & Eating lunch  & 21,798\\
Writing & 7 & Writing on a paper  & 21,434\\
Using laptop & 8 & Using laptop normally  & 22,009\\
Washing face & 9 & Washing face standing  & 22,027\\
Washing hand & 10 & Washing hands standing  & 22,009\\
Swiping &11 & Swiping a surface walking and standing  & 19,186\\
Vacuuming & 12 & Vacuuming a surface while walking and standing  & 22,507\\
Dusting a surface & 13 & Dusting a surface sitting  & 21,513 \\
Brushing Teeth & 14 & Brushing teeth standing & 22,495\\
\bottomrule
\end{tabularx}
\end{threeparttable}
\end{table*}

\subsection{Federated Learning Testbed}

We built a testbed to support training and testing of the proposed HAR classifier in a federated setting (i.e., the TransFed framework). The testbed consists of one aggregation server (master node) and five client (edge) devices. Each client trains its local model for $e=100$ epochs on the local dataset, updates the local model and sends the model back to the aggregation server. When all clients perform $e$ number of epochs, the master node updates the global model and sends it again to the client workers. The process continues in $r$ number of communication rounds. Algorithm~\ref{algo:algorithm} defines the whole implementation of the federated learning using TensorFlow and socket communication. Each client executes Lines 3-5. Whereas, the rest of the algorithm is executed by the aggregation server. In Algorithm~\ref{algo:algorithm}, \texttt{train()} refers to TensorFlow's training (fit) function, and \texttt{send()} refers to the send function of the MLSocket library in Python.

\begin{algorithm}[!ht]
\caption{The federated learning algorithm for training our proposed transformer-based HAR classifier.}
\label{algo:algorithm}
\DontPrintSemicolon

\KwInput{$\mathbf{GM}^r$ -- the global transformer model for the $r$-th round, $\mathbf{LM}_k^r$ -- the local model on the $k$-th client for the $r$-th round, $n$ -- the number of data observations across all clients, $n_k$ -- the number of observations on each client (edge), $\mathbf{LD}_k$ -- the set of local datasets for training on each client, $r$ -- the index of the current round, $e$ -- the number of training epochs per one round, $b$ -- batch size of training data, $\Delta W_k^r$ -- parameters of client $k$ at $r$-th round, $K$ -- the number of clients participating in federated learning.} 
\KwOutput{Trained aggregated and updated model}

\While{while $r \neq 0$}{
\For{each cleint $k$}{
$\mathbf{LM}_k^r=\mathbf{GM}^r$\\
$\mathbf{LM}_k^{r+1}=\text{Train}(\mathbf{LM}_k^r,\mathbf{LD}_k,e,b)$\\
send ($\Delta W_k^{r+1}$)
}
$\mathbf{GM}^{r+1}=\sum_{k=1}^K\frac{n_k}{n}\mathbf{LM}_k^{r+1}$
}
\end{algorithm}

To simulate more realistic scenarios, we used the SSL (Secure Socket Layer) protocol for secure communications between the server and client devices. The transformer-based classifier was trained locally only on five local Raspberry Pi devices (Pi 5 Model B+ with a 1.4GHz, 64-bit quad-core ARMv8 CPU and 1GB LPDDR2 SDRAM) as edge devices. Furthermore, a workstation with an Intel core i-6700HQ CPU and 32 GB RAM was used as the aggregation server. This hardware setup for clients allowed us to simulate what typical home healthcare systems can provide, in terms of computing resources.

\subsection{Data Partitioning and Distribution}

In order to analyze the performance of the proposed transformer in a federated setting we used skew data derived from a non-IID distribution. To achieve a non-IID data over the clients in a federated learning, we group the data by each user in the dataset and split it among 5 different clients, and selected one activity on each client to have 40-50\% less samples. For example, Client 1 has 40-50\% less samples of standing activity compared to others and Client 2 has 40-50\% less samples of sitting activity compared to others.

\subsection{Centralized Setting}

In addition to testing the performance of our proposed TransFed framework based on the proposed transformer, we also tested the proposed transformer in a centralized setting, which demonstrated that it is a general technique that can work under both centralized and federated settings. We tested the proposed transformers with both our new dataset and the well-known public WISDM dataset~\citep{kwapisz2011}, and compared its performance against other state-of-the-art methods. Since most of the existing work used WISDM dataset in centralized setting, testing the proposed transformer-based classifier using public dataset in a centralized setting gives a fair comparison. The details of WISDM can be found in Table~\ref{tab:wisdm-dataset}.

\begin{table}[!ht]
\centering
\caption{Basic information of the activity classes in our collected dataset.}
\label{tab:wisdm-dataset}
\begin{tabularx}{\linewidth}{ccY}
\toprule
Class Name & Class ID & Number of Samples\\
\midrule
Walking & 0 & 424,400\\
Jogging & 1 & 342,177\\
Going Upstairs & 2 & 122,869\\
Going Downstairs & 3 & 3,100,427\\
Sitting & 4 & 59,939\\
Standing & 5 & 48,395\\
\bottomrule
\end{tabularx}
\end{table}

\section{Performance Analysis}
\label{section:performance}

In this section, we provide the performance analysis of the proposed framework based on the above-mentioned experimental setup.

\subsection{Performance Metrics Used}

To measure the classification performance of the proposed transformer based classifier, we use the following four classification metrics widely used for evaluating machine learning models.

\begin{enumerate}
\item \textbf{Accuracy} is defined as the correctly predicted observation divided by total observations, as given below:
\begin{equation}
\text{Accuracy} = \frac{\text{TP}+\text{TN}}{\text{TP}+\text{FP}+\text{FN}+\text{TN}},
\end{equation}
where TP, TN, FP and FN are true positives, true negatives, false positives, and false negatives number of samples, respectively.

\item \textbf{Precision} defined as the number of true positives divided by the total number true positives plus total number of false positives for a given class, given as follow:
\begin{equation}
\text{Precision} = \frac{\text{TP}}{\text{TP}+\text{FP}}.
\end{equation}

\item \textbf{Recall} defined as the total number of true positives divided by total true positives plus false negatives for a given class, given as follow:
\begin{equation}
\text{Recall} = \frac{\text{TP}}{\text{TP}+\text{FN}}.
\end{equation}

\item \textbf{F1-score} is defined as the weighted average of precision and recall, given as follows:

\begin{equation}
\text{F1-Score} = \frac{2 \times \text{Recall} \times \text{Precision}}{\text{Recall} + \text{Precision}}.
\end{equation}
\end{enumerate}

All the above four performance metrics used are defined for binary classifiers only, so for multi-class HAR classifiers we used the one-vs-rest strategy to calculate the performance metrics for each class.

\subsection{Accuracy and Loss (Training and validation)}

Accuracy (one-vs-rest accuracy, where we split multi-class classification into binary classification problem per class) and loss (categorical cross-entropy) are often used to measure how a machine learning classifier's performance evolves during the training process. The trend over time can be used to determine whether the model is properly and ideally trained, to detect anomalies in time (such as over- or under-fitting), and to make necessary adjustments.

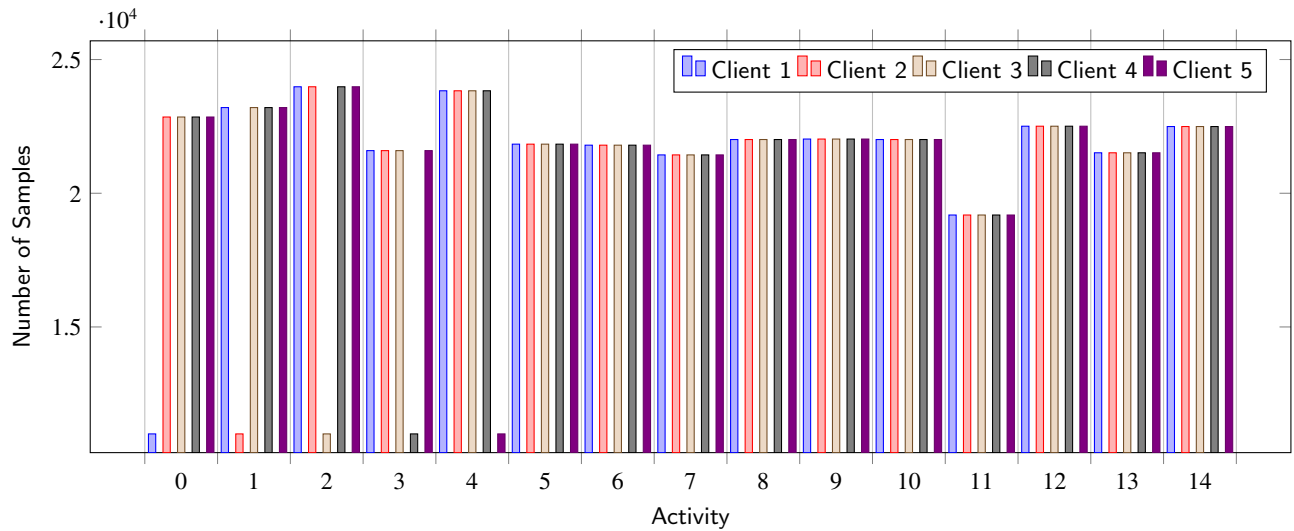
\begin{figure*}
\begin{tikzpicture}
\begin{axis}[
    width=\linewidth,
    height=7cm,
    ymax=25000,
	x tick label style={
		/pgf/number format/1000 sep=},
	ylabel=Number of Samples,
	xlabel=Activity,
	enlargelimits=0.05,
	legend style={legend columns=-1},
	ybar interval=0.5,
]
\addplot 
	coordinates {(0,11000) (1,23204)
		 (2,23982) (3,21594)(4,23832)(5,21836)(6,21798)(7,21434)(8,22009)(9,22027)(10,22009)(11,19186)(12,22507)(13,21513)(14,22495)(15,22495)};
\addplot
 coordinates {(0,22851) (1,11000)
		 (2,23982) (3,21594)(4,23832)(5,21836)(6,21798)(7,21434)(8,22009)(9,22027)(10,22009)(11,19186)(12,22507)(13,21513)(14,22495)(15,22495)};
\addplot
 coordinates {(0,22851) (1,23204)
		 (2,11000) (3,21594)(4,23832)(5,21836)(6,21798)(7,21434)(8,22009)(9,22027)(10,22009)(11,19186)(12,22507)(13,21513)(14,22495)(15,22495)};
\addplot
 coordinates {(0,22851) (1,23204)
		 (2,23982) (3,11000)(4,23832)(5,21836)(6,21798)(7,21434)(8,22009)(9,22027)(10,22009)(11,19186)(12,22507)(13,21513)(14,22495)(15,22495)};
\addplot
 coordinates {(0,22851) (1,23204)
		 (2,23982) (3,21594)(4,11000)(5,21836)(6,21798)(7,21434)(8,22009)(9,22027)(10,22009)(11,19186)(12,22507)(13,21513)(14,22495)(15,22495)};
\legend{Client 1, Client 2, Client 3, Client 4, Client 5}
\end{axis}
\end{tikzpicture}
\caption{Data derived from a non-IID quantity distribution among all clients in the federated setting, where each client contains 40-50\% less data of a given class. The x-axis represents the class ID and the y-axis represents the number of samples.}
\label{fig:data-distribution}
\end{figure*}

To evaluate the performance of the proposed transformer in a federated setting, we used non-IID distribution as shown in Figure~\ref{fig:data-distribution}. We trained the transformer at each client for 100 epochs. The hyper parameters used during the training process are given in Table~\ref{tab:hyper-parameters}. Each client used two transformer layers and a learning rate of 0.01 with an Adam optimizer.

\begin{table}[!ht]
\centering
\caption{Hyper-parameters used by each client for federated learning.}
\label{tab:hyper-parameters}
\begin{tabularx}{\linewidth}{YY}
\toprule
Hyper-Parameter & Value\\
\midrule
Learning Rate & 0.01\\
Number of Epochs & 100\\
Batch Size & 30\\
Weight Decay & 0.001\\
Transformer Layers & 2\\
Multi-attention Heads & 5\\
Input shape & 140$\times$9\\
\bottomrule
\end{tabularx}
\end{table}

\begin{figure*}[!ht]
\centering
\begin{subfigure}[b]{0.3\textwidth}
   \centering
   \includegraphics[width=\textwidth]{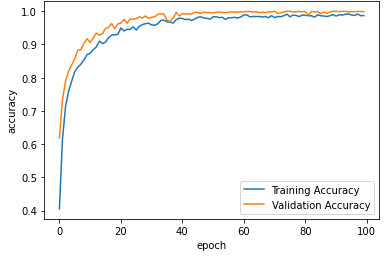}
   \caption{Client 1}
   \label{fig:clinet1_accuracy}
  \end{subfigure}
  \hfill
  \begin{subfigure}[b]{0.3\textwidth}
   \centering
   \includegraphics[width=\textwidth]{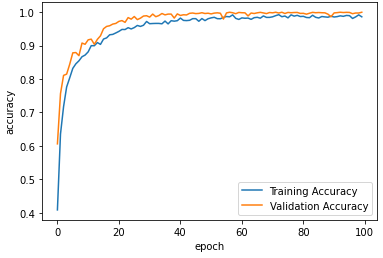}
   \caption{Client 2}
   \label{fig:clinet2_accuracy}
  \end{subfigure}
  \hfill
  \begin{subfigure}[b]{0.3\textwidth}
   \centering
   \includegraphics[width=\textwidth]{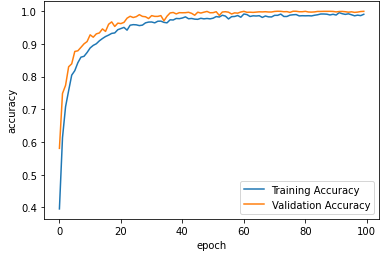}
   \caption{Client 3}
   \label{fig:clinet3_accuracy}
  \end{subfigure}
 \hspace*{\fill}
  \begin{subfigure}[b]{0.3\textwidth}
   \centering
   \includegraphics[width=\textwidth]{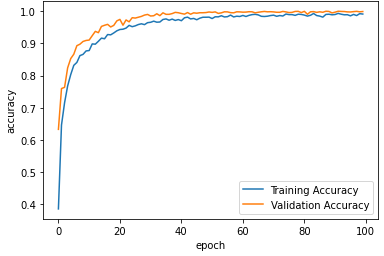}
   \caption{Client 4}
   \label{fig:clinet4_accuracy}
  \end{subfigure}
  \hspace*{\fill}
  \begin{subfigure}[b]{0.3\textwidth}
   \centering
   \includegraphics[width=\textwidth]{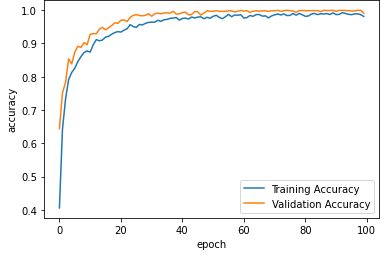}
   \caption{Client 5}
   \label{fig:clinet5_accuracy}
  \end{subfigure}
  \hspace*{\fill}
\caption{Training and validation accuracy of the clients using data derived from a non-IID distribution.}
\label{fig:Accuracy_IID}
\end{figure*}

\begin{figure*}[!ht]
\centering
\begin{subfigure}[b]{0.3\textwidth}
    \centering
    \includegraphics[width=\textwidth]{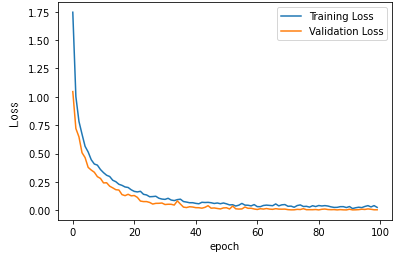}
    \caption{Client 1}
    \label{fig:clinet1_loss}
\end{subfigure}
\quad
 \begin{subfigure}[b]{0.3\textwidth}
    \centering
    \includegraphics[width=\textwidth]{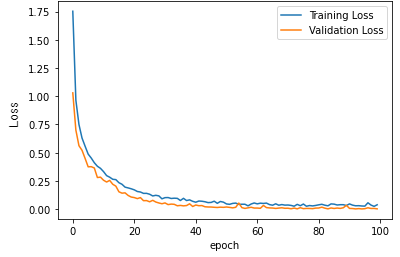}
    \caption{Client 2}
    \label{fig:clinet2_loss}
\end{subfigure}
\quad
\begin{subfigure}[b]{0.3\textwidth}
    \centering
    \includegraphics[width=\textwidth]{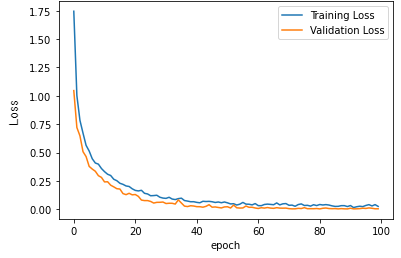}
    \caption{Client 3}
    \label{fig:clinet3_loss}
\end{subfigure}

\begin{subfigure}[b]{0.3\textwidth}
    \centering
    \includegraphics[width=\textwidth]{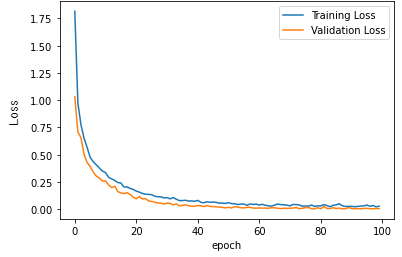}
    \caption{Client 4}
    \label{fig:clinet4_loss}
\end{subfigure}
\quad
\begin{subfigure}[b]{0.3\textwidth}
    \centering
    \includegraphics[width=\textwidth]{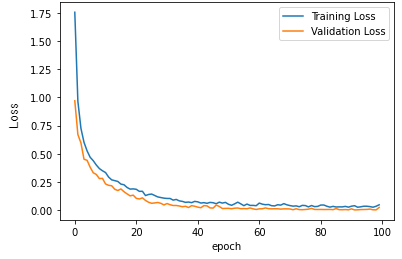}
    \caption{Client 5}
    \label{fig:clinet5_loss}
\end{subfigure}
\caption{Training and validation loss of the clients using data derived from a non-IID distribution.}
\label{fig:Loss_IID}
\end{figure*}

Figures~\ref{fig:Accuracy_IID} and \ref{fig:Loss_IID} show the accuracy and loss curves for each local model at the corresponding local client, respectively. For comparison purposes, we performed 100 iterations (epochs) for each model and draw a point every 10 iterations when drawing, making the curves clearer but still reflecting the overall trend. Among the local models, the performance is almost similar for the non-IID dataset, which indicates that the proposed transformer is robust against imbalanced data caused by a non-IID distribution. Overall, each local model was able to achieve more then 98 percent training and validation accuracy using the non-IID dataset.

\begin{figure}[!ht]
\centering
\begin{subfigure}[b]{0.3\textwidth}
    \centering
    \includegraphics[width=\textwidth]{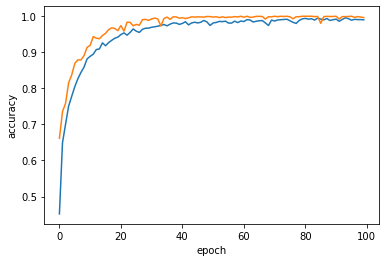}
    \caption{Accuracy of the centralized model with the WISDM dataset}
    \label{fig:cent_wsdm_accuracy}
\end{subfigure}
\hfill
\begin{subfigure}[b]{0.3\textwidth}
    \centering
    \includegraphics[width=\textwidth]{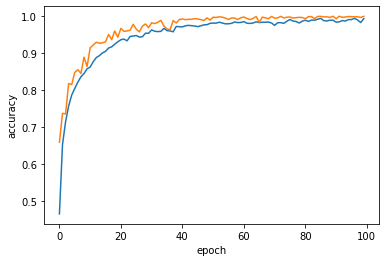}
    \caption{Accuracy of the centralized model with our collected dataset}
    \label{fig:cent_collecteddata_accuracy}
\end{subfigure}
\caption{Training and validation accuracy of the classifier based on the proposed transformer in the centralized setting.}
\label{fig:Accuracy_IID_centralised}
\end{figure}

\begin{figure}[!ht]
\centering
\begin{subfigure}[b]{0.3\textwidth}
   \centering
   \includegraphics[width=\textwidth]{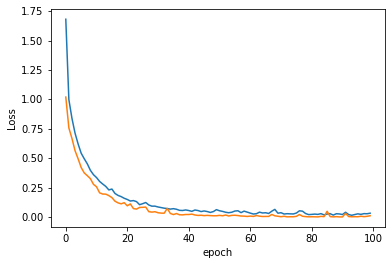}
   \caption{Loss of the centralized model with the WISDM dataset}
   \label{fig:cent_wsdm_loss}
\end{subfigure}
\hfill
\begin{subfigure}[b]{0.3\textwidth}
   \centering
   \includegraphics[width=\textwidth]{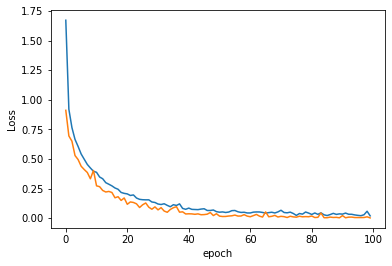}
   \caption{Loss of the centralized model with our collected dataset}
   \label{fig:cent_collecteddata_loss}
\end{subfigure}
\caption{Training and validation loss of the classifier based on the proposed transformer in the centralized setting.}
\label{fig:loss_centralised}
\end{figure}

For the centralized setting, we trained the proposed model using the public WISDM dataset as well as our collected dataset. The hyper-parameters were kept the same as mentioned earlier for the federated setting. Figures~\ref{fig:cent_wsdm_accuracy} and \ref{fig:cent_collecteddata_accuracy} present training and validation accuracy of the proposed transformer in the centralized setting using the WISDM dataset and our collected dataset, respectively. Figures~\ref{fig:cent_wsdm_loss} and \ref{fig:cent_collecteddata_loss} present training and validation loss of the classifier based on the proposed transformer in the centralized setting, using the WISDM dataset and our collected dataset, respectively.

\subsection{Classification Performance}

Tables~\ref{tab:classification_performance1}--\ref{tab:classification_performance5} present classification results obtained on all the five clients using their local non-IID data in the federated setting. Table~\ref{tab:classification_per_FL} shows the classification performance of the global model after performing federated averaging. We tested the global model using a test dataset that was not used to train any of the client models and the proposed transformer achieved an overall accuracy of 98.74\%.

Moreover, Tables~\ref{tab:classification_wisdm} and \ref{tab:classification_balanced_collected_data} present the classification performance of the proposed transformer transformer based classifier in the centralized setting, using the WISDM dataset and our collected dataset, respectively. Overall the classifier based on our proposed transformer achieved an accuracy of 99.14\% and 98.89\% with our collected dataset and the WISDM dataset, respectively. Since the WISDM dataset has much more unbalanced class samples than our collected dataset, it is not surprising to see the performance is (slightly) lower compared with our collected dataset, since imbalanced data are harder to learn.

\begin{table}[!ht]
\centering
\caption{Classification performance of the clients using data derived from a non-IID distribution: Client 1 with 50\% less ``Standing'' data.}
\label{tab:classification_performance1}
\begin{tabularx}{\linewidth}{cYYY}
\toprule
Activity & Precision & Recall & F1-score\\
\midrule
Standing & 100\% & 100\% & 100\%\\
Sitting & 100\% & 100\% & 100\%\\
Walking & 100\% & 100\% & 100\%\\
Jogging & 100\% & 100\% & 100\%\\
Going Upstairs & 100\% & 100\% & 100\%\\
Going Downstairs & 100\% & 100\% & 100\%\\
Eating & 100\% & 100\% & 100\%\\
Writing & 100\% & 100\% & 100\%\\
Using Laptop & 100\% & 100\% & 100\%\\
Washing Face & 100\% & 99.0\% & 99\%\\
Washing Hand & 99.0\% & 100\% & 99.0\%\\
Swiping & 95.0\% & 100\% & 98.0\%\\
Vacuuming & 100\% & 96.0\% & 98.0\%\\
Dusting & 100\% & 98.0\% & 99.0\%\\
Brushing Teeth & 100\% & 100\% & 100\%\\
\bottomrule
\end{tabularx}
\end{table}

\begin{table}[!ht]
\centering
\caption{Classification performance of the clients using data derived from a non-IID distribution: Client 2 with 50\% less ``Sitting'' data.}
\label{tab:classification_performance2}
\begin{tabularx}{\linewidth}{cYYY}
\toprule
Activity & Precision & Recall & F1-score\\
\midrule
Standing & 100\% & 100\% & 100\%\\
Sitting &  98.0\% & 100\% & 99.0\%\\
Walking &  100\% & 98.0\% & 99.0\%\\
Jogging &  99.0\% & 99.0\% & 99.0\%\\
Going Upstairs &  98.0\% & 96.0\% & 97.0\%\\
Going Downstairs & 95.0\% & 99.0\% & 97.0\%\\
Eating & 98.0\% & 98.0\% & 97.0\%\\
Writing & 100\% & 96.0\% & 99.0\%\\
Using Laptop & 96.0\% & 100\% & 98.0\%\\
Washing Face & 99.0\% & 100\% & 99.0\%\\
Washing Hand & 97.0\% & 97.0\% & 97.0\%\\
Swiping & 96.0\% & 95.0\% & 96.0\%\\
Vacuuming & 93.0\% & 94.0\% & 93.0\%\\
Dusting & 96.0\% & 92.0\% & 94.0\%\\
Brushing Teeth & 97.0\% & 98.0\% & 97.0\%\\
\bottomrule
\end{tabularx}
\end{table}

\begin{table}[!ht]
\centering
\caption{Classification performance of the clients using data derived from a non-IID distribution: Client 3 with 50\% less ``Walking'' data.}
\label{tab:classification_performance3}
\begin{tabularx}{\linewidth}{cYYY}
\toprule
Activity & Precision & Recall & F1-score\\
\midrule
Standing & 100\%  & 100\% & 100\%\\
Sitting & 100\% & 100\% & 100\%\\
Walking & 99.0\% & 100\% & 100\%\\
Jogging & 98.0\% & 98.0\% & 98.0\%\\
Going Upstairs & 98.0\% & 97.0\% & 98.0\%\\
Going Downstairs & 98.0\% & 98.0\% & 98.0\%\\
Eating & 95.0\% & 97.0\% & 96.0\%\\
Writing & 99.0\% & 98.0\% & 99.0\%\\
Using Laptop & 96.0\%  & 97.0\% & 96.0\%\\
Washing Face & 96.0\% & 99.0\% & 97\%\\
Washing Hand & 98.0\% & 95.0\% & 96.0\%\\
Swiping & 96.0\% & 92.0\% & 94.0\%\\
Vacuuming & 95.0\% & 97.0\% & 96.0\%\\
Dusting & 96.0\% & 95.0\% & 96.0\%\\
Brushing Teeth & 97.0\% & 98.0\% & 98.0\%\\
\bottomrule
\end{tabularx}
\end{table}

\begin{table}[!ht]
\centering
\caption{Classification performance of the clients using data derived from a non-IID distribution: Client 4 with 50\% less ``Jogging'' data.}
\label{tab:classification_performance4}
\begin{tabularx}{\linewidth}{cYYY}
\toprule
Activity & Precision & Recall & F1-score\\
\midrule
Standing & 100\% & 100\% & 100\%\\
Sitting & 100\% & 97.0\% & 99.0\%\\
Walking & 97.0\% & 100\% & 99.0\%\\
Jogging & 99.0\% & 100\% & 100\%\\
Going Upstairs & 99.0\% & 98.0\% & 98.0\%\\
Going Downstairs & 98.0\% & 99.0\% & 98.0\%\\
Eating & 97.0\% & 97.0\% & 97.0\%\\
Writing & 99.0\% & 98.0\% & 98.0\%\\
Using Laptop & 97.0\% & 99.0\% & 98.0\%\\
Washing Face & 98.0\% & 97.0\% & 98.0\%\\
Washing Hand & 96.0\% & 98.0\% & 97.0\%\\
Swiping & 96.0\% & 94.0\% & 95.0\%\\
Vacuuming & 93.0\% & 94.0\% & 94.0\%\\
Dusting & 95.0\% & 94.0\% & 95.0\%\\
Brushing Teeth & 99.0\% & 98.0\% & 98.0\%\\
\bottomrule
\end{tabularx}
\end{table}

\begin{table}[!ht]
\centering
\caption{Classification performance of the clients using data derived from a non-IID distribution: Client 5 with 50\% less ``Going Upstairs'' data.}
\label{tab:classification_performance5}
\begin{tabularx}{\linewidth}{cYYY}
\toprule
Activity & Precision & Recall & F1-score\\
\midrule
Standing & 100\% & 100\% & 100\%\\
Sitting & 99.0\% & 97.0\% & 98.0\%\\
Walking & 97.0\% & 98.0\% & 98.0\%\\
Jogging & 99.0\% & 100\% & 99.0\%\\
Going Upstairs & 100\% & 100\% & 100\%\\
Going Downstairs & 99.0\% & 99.0\% & 99.0\%\\
Eating & 96.0\% & 96.0\% & 96.0\%\\
Writing & 100\% & 97.0\% & 98.0\%\\
Using Laptop & 96.0\% & 100\% & 98.0\%\\
Washing Face & 97.0\% & 97.0\% & 97.0\%\\
Washing Hand & 95.0\% & 95.0\% & 95.0\%\\
Swiping & 94.0\% & 95.0\% & 95.0\%\\
Vacuuming & 97.0\% & 96.0\% & 96.0\%\\
Dusting & 96.0\% & 97.0\% & 96.0\%\\
Brushing Teeth & 99.0\% & 98.0\% & 98.0\%\\
\bottomrule
\end{tabularx}
\end{table}

\begin{table}[ht!]
\centering
\caption{Classification performance of the global model after federated averaging using data derived from a non-IID distribution.} \label{tab:classification_per_FL}
\begin{tabularx}{\linewidth}{cYYY}
\toprule
Activity & Precision & Recall & F1-score\\
\midrule
Standing & 100\% & 100\% & 100\%\\
Sitting & 100\% & 100\% & 100\%\\
Walking & 99.0\% & 100\% & 100\%\\
Jogging & 99.0\% & 99.0\% & 99.0\%\\
Going Upstairs & 96.0\% & 94.0\% & 95.0\%\\
Going Downstairs & 97.0\% & 97.0\% & 97.0\%\\
Eating & 99.0\% & 97.0\% & 98.0\%\\
Writing & 100\% & 99.0\% & 100\%\\
Using Laptop & 99.0\% & 100\% & 99.0\%\\
Washing Face & 100\% & 98.0\% & 99.0\%\\
Washing Hand & 93.0\% & 100\% & 97.0\%\\
Swiping & 97.0\% & 90.0\% & 94.0\%\\
Vacuuming & 90.0\% & 98.0\% & 94.0\%\\
Dusting & 99.0\% & 94.0\% & 97.0\%\\
Brushing Teeth & 96.0\% & 99.0\% & 97.0\%\\
\bottomrule
\end{tabularx}
\end{table}

\begin{table}[ht!]
\centering
\caption{Classification performance of the proposed transformer in the centralized setting using WISDM dataset~\citep{kwapisz2011}.} \label{tab:classification_wisdm}
\begin{tabularx}{\linewidth}{cYYY}
\toprule
Activity & Precision & Recall & F1-score\\
\midrule
Walking & 100\% & 100\% & 100\%\\
Jogging & 100\% & 100\% & 100\%\\
Going Upstairs & 95.0\% & 96.0\% & 96.0\%\\
Going Downstairs & 98.0\% & 99.0\% & 99.0\%\\
Sitting & 100\% & 100\% & 100\%\\
Standing & 97.0\% & 97.0\% & 97.0\%\\
\bottomrule
\end{tabularx}
\end{table}

\begin{table}[ht!]
\centering
\caption{Classification performance of the proposed transformer in the centralized setting using our collected dataset.} \label{tab:classification_balanced_collected_data}
\begin{tabularx}{\linewidth}{cYYY}
\toprule
Activity & Precision & Recall & F1-score\\
\midrule
Standing & 100\% & 100\% & 100\%\\
Sitting & 100\% & 99.0\% & 99.0\%\\
Walking & 99.0\% & 99.0\% & 99.0\%\\
Jogging & 98.0\% & 98.0\% & 98.0\%\\
Going Upstairs & 99.0\% & 95.0\% & 97.0\%\\
Going Downstairs & 97.0\% & 99.0\% & 98.0\%\\
Eating & 99.0\% & 98.0\% & 98.0\%\\
Writing & 98.0\% & 99.0\% & 98.0\%\\
Using Laptop & 97.0\% & 99.0\% & 98.0\%\\
Washing Face & 100\% & 97.0\% & 99.0\%\\
Washing Hand & 96.0\% & 100\% & 98.0\%\\
Swiping & 95.0\% & 90.0\% & 93.0\%\\
Vacuuming & 93.0\% & 98.0\% & 95.0\%\\
Dusting & 99.0\% & 93.0\% & 96.0\%\\
Brushing Teeth & 96.0\% & 100\% & 98.0\%\\
\bottomrule
\end{tabularx}
\end{table}

\subsection{Confusion Matrices}

\begin{figure*}[!ht]
\centering
\includegraphics[width=0.75\linewidth]{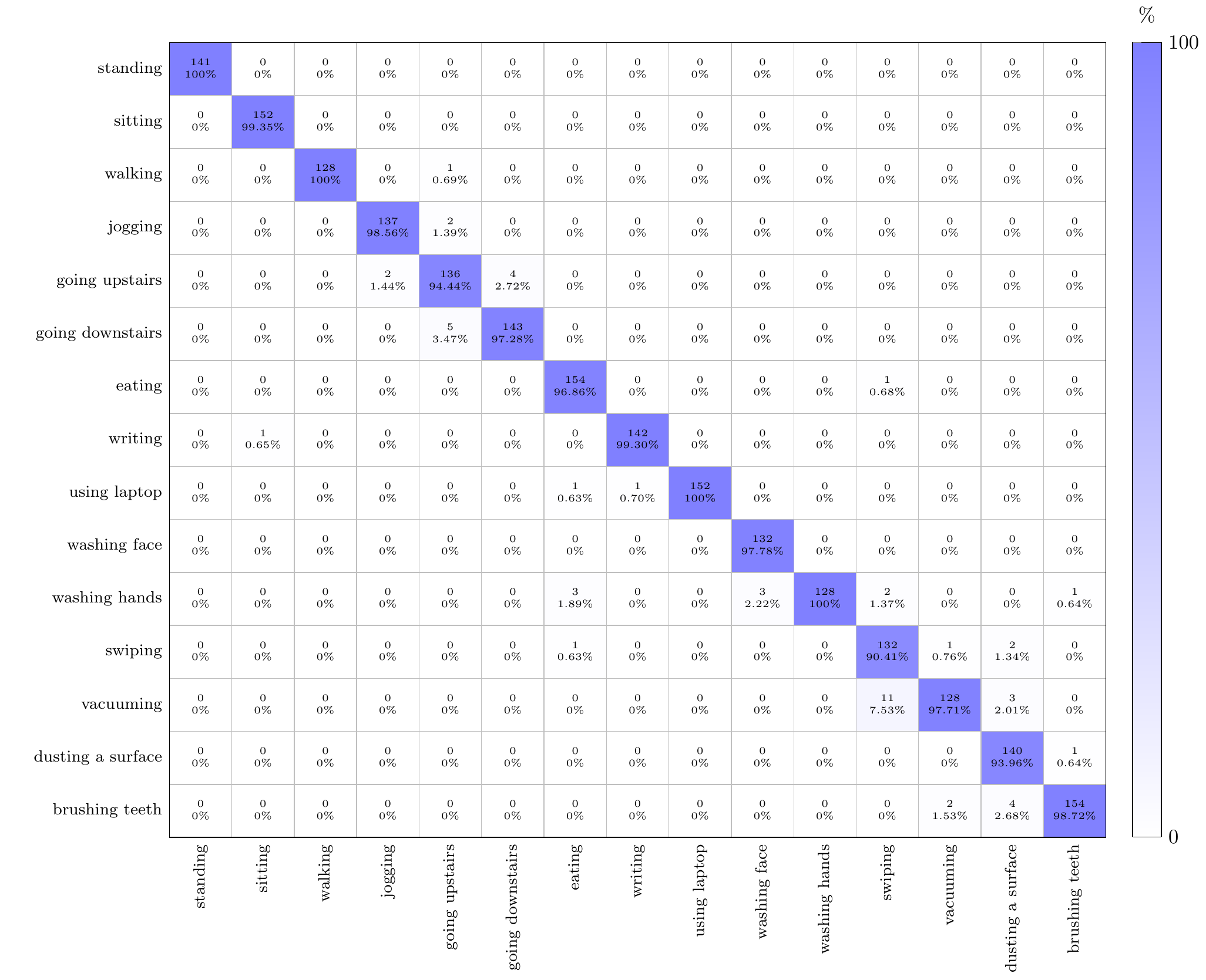}
\caption{The confusion matrix obtained with the final global transformer model in a federated setting with five clients, using data derived from a non-IID distribution.}
\label{fig:CM_fl_unbalanced_IID}
\end{figure*}

\begin{figure*}[!ht]
\centering
\includegraphics[width=0.75\linewidth]{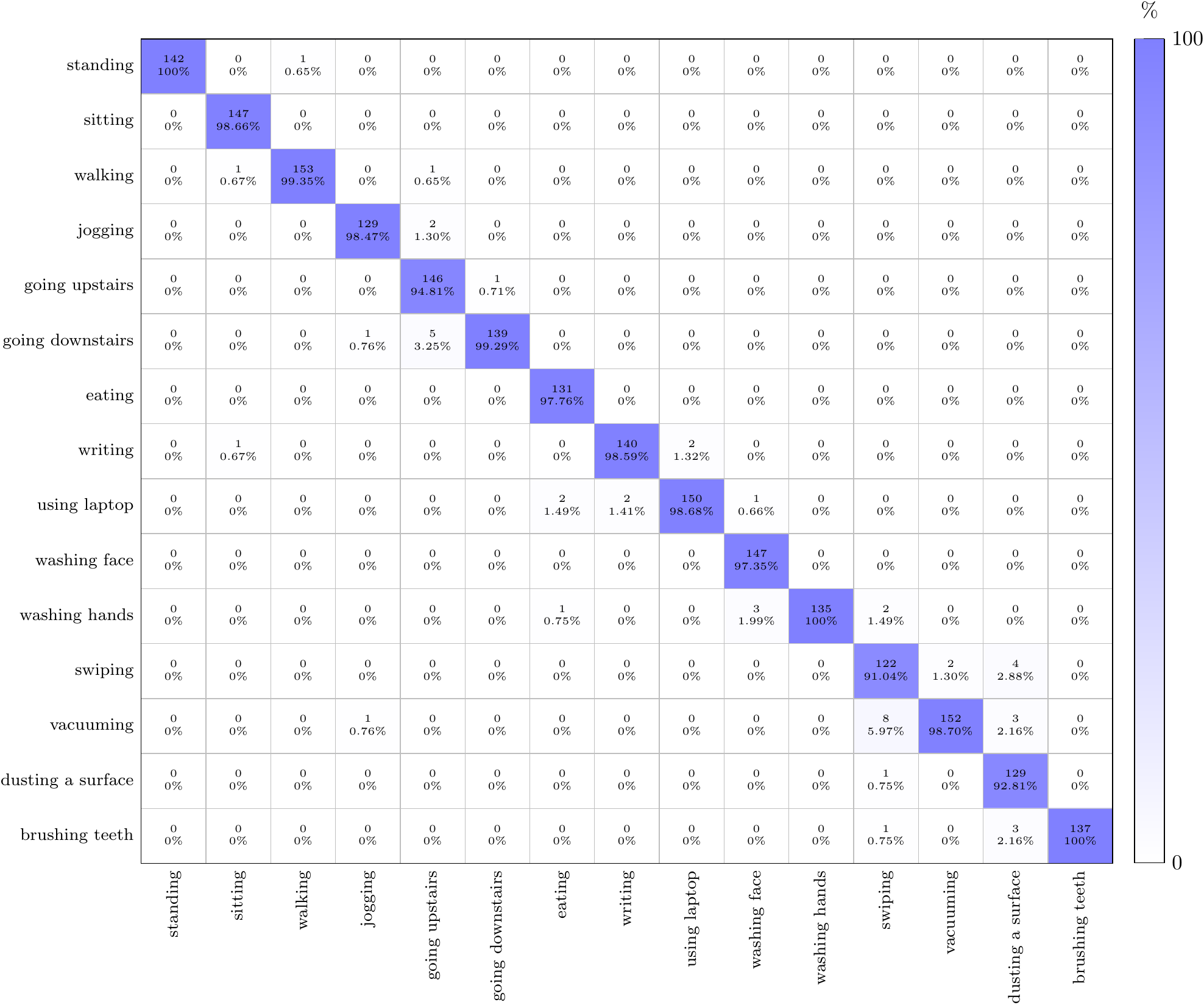}
\caption{The confusion matrix obtained with the transformer model in a centralized setting using 5-fold cross-validation, using the balanced collected data.}
\label{fig:CM_centralized}
\end{figure*}

\begin{figure}[!ht]
\centering
\includegraphics[width=\linewidth]{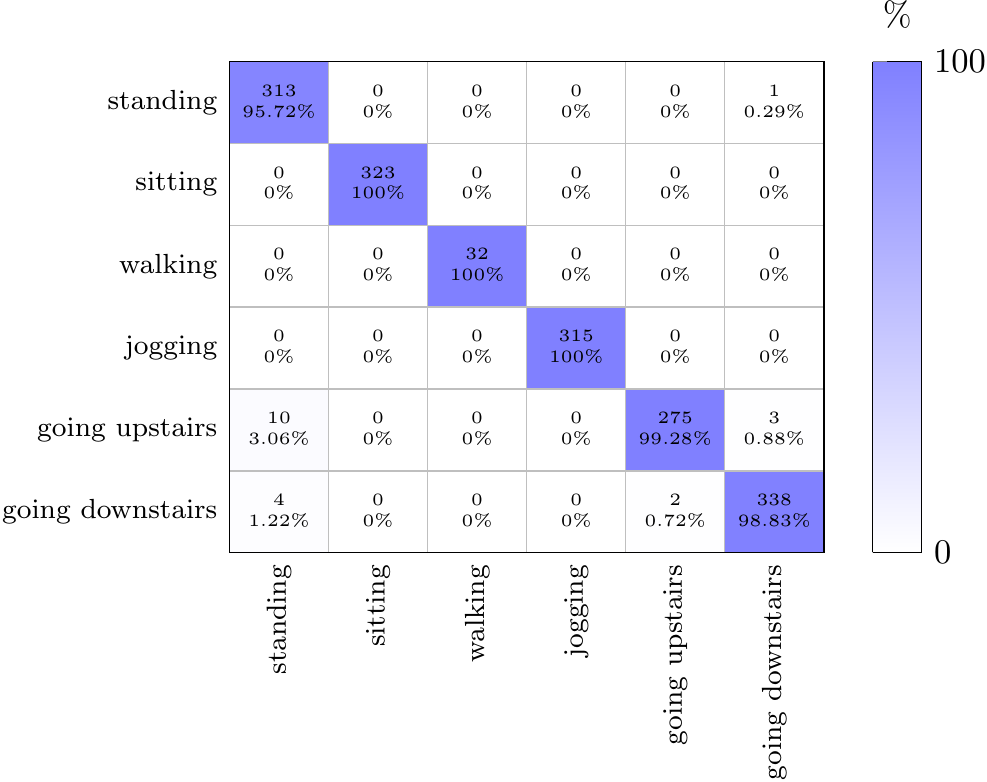}
\caption{The confusion matrix obtained with the transformer model in the centralized setting , using the WISDM dataset.}
\label{fig:WISDM_CM_centralized}
\end{figure}

A confusion matrix, also known as an error matrix, is an $n\times n$ matrix or table that shows how each class is classified into all the $n$ classes of a classifier. The diagonal elements of a confusion matrix shows the correct classification results and other cells show different misclassification rates. Hence, we evaluate the proposed transformer using confusion matrix to determine where exactly the transformer miss-classifies the classes during testing. Figure~\ref{fig:CM_fl_unbalanced_IID} presents the confusion matrix obtained with the updated global model in the federated setting, using our collected dataset. Figure~\ref{fig:CM_centralized} presents the confusion matrix obtained using the classifier based on our proposed transformer in the centralized setting, using our collected dataset. In both figures, the x-axis indicates the predicted class labels and the y-axis indicates the ground true class labels. We can see that in both settings the proposed HAR classifier worked very well with similar performance across all the 15 classes. Similarly, Figure~\ref{fig:WISDM_CM_centralized} presents the confusion matrix obtained using the WISDM dataset in the centralized settings. It can be seen that the proposed transformer achieved almost perfect classification results for all classification activities.

From the confusion matrices, it can be observed that mis-classifications occurred more between activities that involve similar body movements, e.g., swiping and vacuuming. It can also be observed that, even for these similar but different human activities, the proposed transformer achieved a very good performance.

\subsection{Comparison with State-of-the-Art Methods}

In this subsection, we compare the proposed transformer with existing state-of-the-art methods. Table~\ref{tab:Comparision_computation} compares two key features of our proposed transformers with methods based on RNNs and CNNs. RNN-based models do not allow parallelization during training because of their sequential nature, which makes the model computationally slow and expensive. CNN-based methods can perform parallel computation, but they are computationally expensive because of the convolution function. Our new method based on the proposed transformer completely eliminates recurrence and convolution and replaces them with a self-attention mechanism to establish dependencies between the input and the output. It is the first type of architecture to rely entirely on attention to calculate representations of the input and the output. In addition, transformers leave more room for parallelization. RNNs and CNNs use a large number of parameters (usually hundreds of thousands or even more), but the proposed transformer only uses 14,697 parameters. Moreover, unlike traditional transformers, the proposed transformer uses a single patch instead of using multiple-patches. Therefore, the proposed transformer is also much more computationally efficient.

\begin{table}[!ht]
\centering
\caption{Comparison of our transformer-based approach with those based on RNNs and CNNs, in terms of computation costs.}
\label{tab:Comparision_computation}
\begin{tabularx}{\linewidth}{YYY}
\toprule
Method & Parallelization & Computationally\\
 & & Expensive\\
\midrule
RNNs & No  & Yes\\
CNNs & Yes & Yes\\
Transformers & Yes & No\\
\bottomrule
\end{tabularx}
\end{table}

We also compared the performance of TransFed global model and the centralized classifier based on the proposed transform with those of selected state-of-the-art HAR methods in the literature~\citep{sozinov2018human, ignatov2018real, murad2017deep}, including five working in a centralized setting and one in the federated setting. These three state-of-the-art methods were chosen because their performance results were reported using the WISDM dataset, which allow a direct comparison of the performance results. Table~\ref{tab:performance_comparision} shows the comparison results with the six selected state-of-the-art methods. It is obvious that our proposed methods achieved a substantial improvement in terms of accuracy, when trained and tested using the our new dataset and the WISDM dataset. Specifically, our method achieved an accuracy of 98.74\% and 99.14\% in the federated and centralized settings, respectively, using our collected dataset. Furthermore, using the WISDM dataset in the centralized setting, it achieved an overall accuracy of 98.89\%. Hence, from Table~\ref{tab:performance_comparision} it can be seen that the proposed transformer outperforms existing state-of-the-art methods, in both centralized and federated settings.

\begin{table}[!ht]
\centering
\begin{threeparttable}
\caption{Comparison with selected state-of-the-art methods for HAR classification.}
\label{tab:performance_comparision}
\begin{tabularx}{\linewidth}{cYYY}
\toprule 
Scheme & Centralized or Federated & Number of Activities & Accuracy\\
\midrule
\citep{sozinov2018human}\tnote{a} & Federated & 6 & 89.00\%\\
\citep{ignatov2018real}\tnote{a} & Centralized & 6 & 97.63\%\\ 
\citep{murad2017deep}\tnote{a} & Centralized & 6 & 96.70\%\\
Proposed\tnote{a} & Centralized & 6 & 98.89\%\\
Proposed\tnote{b} & Federated & 15 & 98.74\%\\
Proposed\tnote{b} & Centralized & 15 & 99.14\%\\
\bottomrule
\end{tabularx}
\begin{tablenotes}
\small
\item[a] Using the WISDM dataset.

\item[b] Using our new dataset.
\end{tablenotes}
\end{threeparttable}
\end{table}

\section{Further Discussions}
\label{section:discussions}

In summary, our experimental results showed that our proposed method is a step forward to protect the privacy of users using transformer-based FL, which uses very few parameters while providing high accuracy. However, although FL can provide better privacy protection of local data against the global server, various security and privacy attacks have been proposed to make simple FL architecture less secure and privacy-preserving~\citep{hathaliya2020exhaustive}, therefore, one future research direction is to investigate how the proposed method can be further enhanced to be robust against known security and privacy attacks.

In addition to further improve data security and privacy, a major limitation of our work is that our newly constructed dataset was based on only five human participants and a more artificially constructed home care scenario. It is therefore important to evaluate the proposed work in more realistic home healthcare settings, and ultimately move the proposed HAR classification method into real-world usage.
For such future work, involvement of patients, carers and health professionals is vital in all stages of the research process, following well-established standard procedures and guidelines such as the UK Standards for Public Involvement~\citep{UK_standards_public_involvement_2016}.
One important aspect of the real-world-facing research is to carefully evaluate the acceptability and usability of body sensors used to ensure that they are the right ones for the target patients. This suggests that different sets of body sensors may have to be used for patients with different conditions or preferences, so we need to investigate how the proposed HAR classifier will change w.r.t.~the different sets of sensors. These include scenarios where no body sensors can be put on the body of a patient, so computer vision based approaches relying on monitoring cameras and microphones will need investigation, which will involve very different machine learning models from those we used for the proposed work in this paper.

\section{Conclusions}

In this paper, we proposed a novel single-patch lightweight transformer for HAR. We examined the use of transformers as a HAR classifier. The purpose of the lightweight transformer was to provide a state-of-the-art classification performance while keeping it computationally efficient. For our proposed transformer-based HAR classifier, we examined its performance in both federated and centralized settings, under a non-IID data distribution. To test the performance of the proposed transformer in the federated setting, we developed a framework called TransFed and designed a testbed to collect data from five human participants who conducted 15 different activities in a simulated home environment.

Our extensive experimental results confirmed that the proposed transformer can provide better performance compared with a number of state-of-the-art CNN- and RNN-based HAR classifiers, while providing a standardized and automated way to accomplish the feature learning step. Furthermore, the federated setting used by our proposed framework TransFed can help improve data privacy, which is a major issue in centralized approaches.

\section*{Acknowledgments}

This research work was supported by the I-SITE Université Lille Nord-Europe 2021 of France under grant No.~I-COTKEN-20-001-TRAN-RAZA.

\printcredits

\bibliographystyle{elsarticle-num}
\bibliography{main.bib}

\end{document}